\documentclass[conference]{IEEEtran}
\IEEEoverridecommandlockouts
\usepackage{cite}
\usepackage{amsmath,amssymb,amsfonts}
\usepackage{algorithmic}
\usepackage{graphicx}
\usepackage{textcomp}
\usepackage{xcolor}
\usepackage{subcaption}

\def\BibTeX{{\rm B\kern-.05em{\sc i\kern-.025em b}\kern-.08em
    T\kern-.1667em\lower.7ex\hbox{E}\kern-.125emX}}
\begin{document}

\title{Using Topological Framework for the Design of  Activation Function and Model Pruning in Deep Neural Networks
}

\author{\IEEEauthorblockN{1\textsuperscript{st} Yogesh Kochar}
\IEEEauthorblockA{\textit{Samsung India Research Bangalore} \\
\textit{name of organization (of Aff.)}\\
Bangalore, India \\
yogesh.kochar@gmail.com}
\and
\IEEEauthorblockN{2\textsuperscript{nd} Sunil Kumar Vengalil}
\IEEEauthorblockA{\textit{International Institute of Information Technology} \\
Bangalore, India \\
vengalilsunilkumar@gmail.com}
\and
\IEEEauthorblockN{3\textsuperscript{rd} Neelam Sinha}
\IEEEauthorblockA{\textit{International Institute of Information Technology} \\
Bangalore, India \\
neelam.sinha@iiitb.ac.in}
}

\maketitle

\begin{abstract}
    Success of deep neural networks in diverse tasks across domains of computer vision, speech recognition and natural language processing, has necessitated understanding the dynamics of training process and also working of trained models.
    Two independent contributions of this paper are 1) Novel activation function for faster training convergence  2) Systematic pruning of filters of models trained irrespective of activation function.
    We analyze the topological transformation of the space of training samples as it gets transformed by each successive layer during training, by changing the activation function.
    The impact of changing activation function on the convergence during training is reported for the task of binary classification.
    A novel activation function aimed at faster convergence for classification tasks is proposed.
    Here, Betti numbers are used to quantify topological complexity of data.
    Results of experiments on popular synthetic binary classification datasets with large Betti numbers(>150) using MLPs are reported.
    Results show that the proposed activation function results in faster convergence requiring fewer epochs by a factor of 1.5  to 2, since Betti numbers reduce faster across layers with the proposed activation function.
    The proposed methodology was verified on benchmark image datasets: fashion MNIST, CIFAR-10 and cat-vs-dog images, using CNNs.
    Based on empirical results, we propose a novel method for pruning a trained model.
    The trained model was pruned by eliminating filters that transform data to a topological space with large Betti numbers.
    All filters with Betti numbers greater than 300 were removed from each layer without significant reduction in accuracy.
    This resulted in faster prediction time and reduced memory size of the model.
\end{abstract}


\section{Introduction}
\label{sec:intro}
Deep neural networks have become the default choice for solving many  complex tasks involving high dimensional datasets in machine learning which were otherwise either partially solved or not solved at all.
However, choosing the right architecture (like the  selection  of hyper parameters activation function, number of layers and number of units per layer)  for a specific task is mostly based on trial and error or based on the previous empirical results.
The transfer function of each layer and the entire neural network is just treated as a complex, unknown and nonlinear function parameterized by weights and biases.

In this study we investigate some of the  desired characteristics a neural network architecture should have for solving a classification task.
We derive our results based on the topology of the space of training data and how this topology changes as data is transformed by each layer.

Topology is a field of mathematics that studies the shape of objects and associated invariances like connectedness, number of holes etc.
It is observed that many real datasets when viewed as point cloud dataset in a high dimensional space follow  certain topology.
For example the study  in \cite{carlsson2009topology} shows that the image patches obtained from natural images follow the topology of a Klein bottle.
Topological data analysis\cite{carlsson2009topology}\cite{chazal2017introduction} uses topological tools (like persistent topology) for analysing point cloud dataset in order to identify and characterize underlying structures in a dataset.

In their recent study, Gregory et.al. \cite{naitzat2020topology}  quantifies topological complexity using Betti numbers.
They observe that topological changes to data across layers of a network remain robust under different instances of training.
They further observe that, compared to smooth activation functions like sigmoid and tanh, non-homeomorphic activation functions like ReLU helps in changing the topology of data faster.

Our work is motivated by \cite{naitzat2020topology} where the transfer function of each layer is looked at, based on  how the layer changes the topology of the data.
Most real world datasets have non-trivial complex topology, and in order to perform classification each layer of the neural network transforms the entire space of data to a simpler topology.
This leads us to the conclusion that in order to achieve classification, each layer of the neural networks should be able to change the topology of data and hence we need a non-homeomorphic transformation at each layer.
This is achieved by activation functions with discontinuity like ReLU. We followed the approach in \cite{naitzat2020topology} and used betti numbers to quantify  topological complexity of the point cloud dataset.

In addition to the above insight that the activation function should be non-homeomorphic, we also hypothesize that a many-to-one transfer function can help to bring samples from the same class closer in the transformed space.
Based on this hypothesis, we introduce a new activation function with multiple many-to-one regions and multiple discontinuities.
Results of our experiments show that, with the proposed activation function the network converges faster as compared to commonly used activation functions like ReLU and sigmoid.
It is also seen that the betti number, computed using persistent homology \cite{naitzat2020topology}, reduces faster with the proposed activation function.

The major contributions of our paper are:
\begin{enumerate}
\item We provide new guidelines for designing  activation functions for supervised classification tasks. We illustrate the guideline by proposing a  new family of activation functions.
\item We propose an easy technique for neural network pruning (i.e reducing the parameters in a trained model without significant decrease in accuracy) using betti numbers computed on the output feature space of each layer.
\end{enumerate}

\begin{figure}[!t]
\centering
  \includegraphics[width=.5\linewidth]{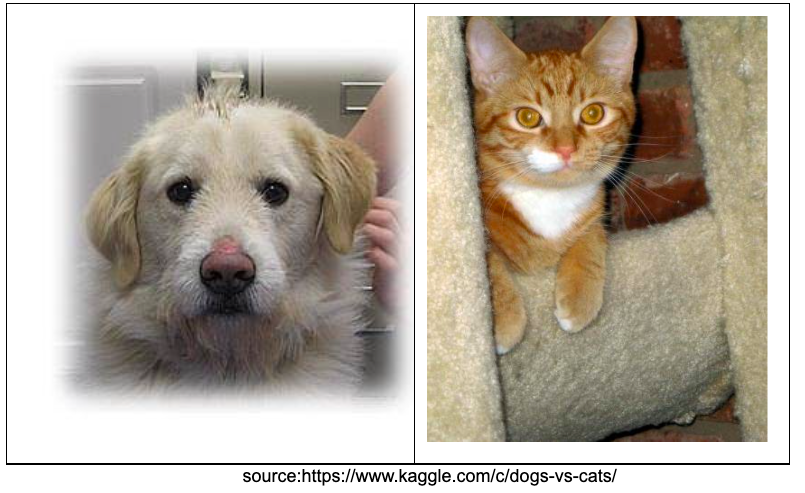}
\vspace{0.1in}
    \caption{Sample images from Cat-Dog dataset}
    \label{catdog}
\end{figure}

\begin{figure}[t!]
\begin{subfigure}{.5\textwidth}
  \centering
  % include first image
  \includegraphics[width=.8\linewidth]{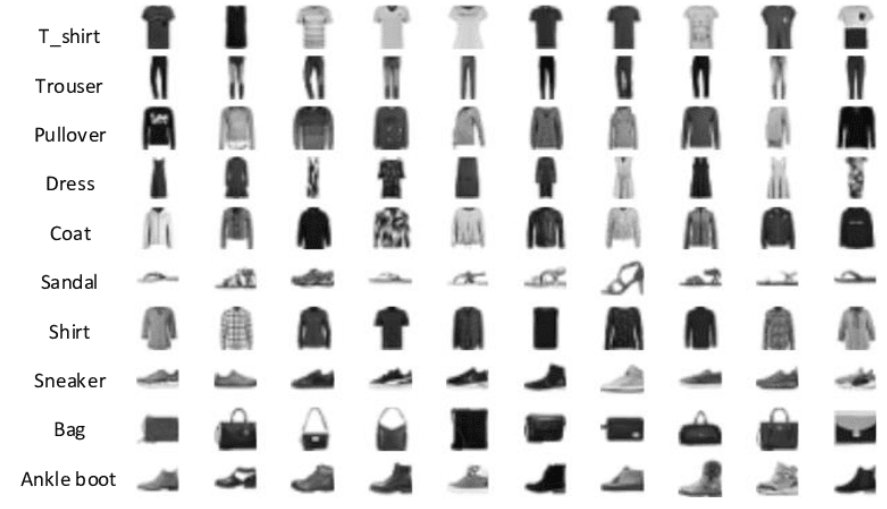}
 % \source{text}
  \caption{Sample images from Fashion Mnist Dataset}
  \caption*{Source:Images taken from \cite{xiao2017fashion}}

  \label{fashionmnist}
\end{subfigure}
\begin{subfigure}{.5\textwidth}
  \centering
  % include second image
  \includegraphics[width=.7\linewidth]{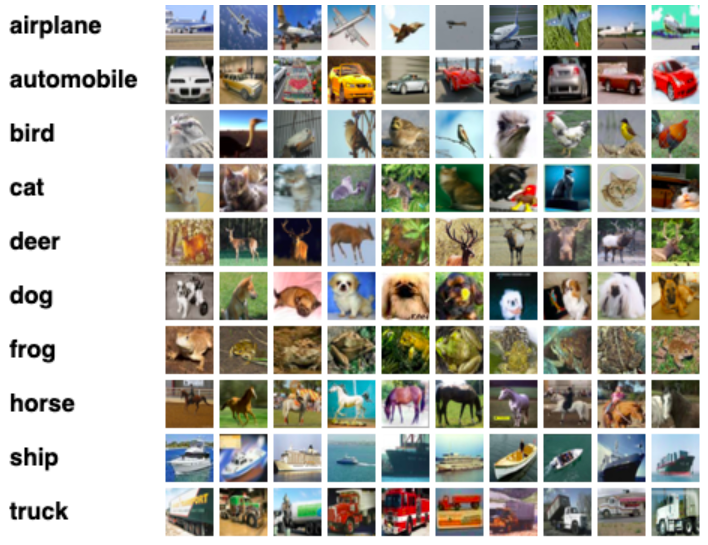}
  \caption{Sample images from Cifar 10 dataset}
  \caption*{Source:https://www.cs.toronto.edu/~kriz/cifar.html}
  \label{cifar10}
\end{subfigure}
\vspace{.1in}
\caption{Sample training images used}

\label{Sample training images used}
\end{figure}

\section{Related Work}
\subsection{Topological Data Analysis}
Topological Data Analysis\cite{chazal2017introduction}\cite{smith2021topological}, is an approach for characterising a dataset using persistent homology \cite{edelsbrunner2008persistent}.
In 2008, Gunnar Carlsson et.al \cite{carlsson2008local} conducted a qualitative study on  $3\times3$ image patches taken from natural images and  results of their study showed that the manifold of  high contrast image patches is homeomorphic to that of Klein bottle.

With the use of deep neural networks for implementing machine learning tasks, like image classification, object detection and segmentation, when the data dimensionality and sample size is huge the challenge of determining the right architecture for a given dataset became a hot area of research interest.
Geometric deep learning, refers to the application of deep neural networks for huge datasets with complex manifold space, not necessarily Euclidean.
The Study in  \cite{bronstein2017geometric} provides an  survey of geometric deeplearning.
 E. Saucan et.al introduces a new sampling technique for sampling manifolds in high dimensional spaces \cite{saucan2007geometric}.

Analysing the dependency between complexity of data and learnability/generalizability of a given network architecture is another interesting area to explore.
For example, the study conducted by William H Guss et.al \cite{chazal2017introduction}  in 2017 brings in the notion of topological capacity of a neural network.
Their empirical results show that learnability of a neural network and  topological complexity of the dataset, computed using persistent homology, are dependent.

\subsection{Activation function and Training convergence}
Since the introduction of ReLU activation function in \cite{krizhevsky2012imagenet} as an alternative for sigmoid and tanh functions, many different variations of it like Leaky ReLU, PReLU, ELU, Threshold ReLU etc has been tried out for faster training convergence and better classification accuracy.
Bounded ReLU activation function was suggested by Shan Sung et. al \cite{liew2016bounded} for better generalizability and training convergence.
In 2017, Ramachandran P et.al. \cite{ramachandran2017searching} used automatic search techniques to look for new activation function.
They evaluated their  best reported activation function $f(x)=x.sigmoid(\beta x)$ on Imagenet using existing best performing architecture and reported 0.9\% improvement on classification accuracy.

Our work differs from all of these as we are using topological simplification as a basis for deriving new activation and we propose that an activation function with many-to-one regions can reduce topological complexity of data.

\subsection{Network Pruning}
Pruning refers to reducing the size of a network (either during training or  for the trained model), by elimination of insignificant parameters,  so that the model becomes compact and can be used on devices with low computing resources.
Pruning can be done at various levels of network architecture like 1) Removing the entire layer 2) Removing filters within a layer,  or 3) Removing individual neurons.

Jian-Hao Luo et. al.\cite{Luo_2017_ICCV}, proposes a framework, which they call ThiNet, for pruning networks at filter level both during training and after training during  inference.
Since the entire filter is removed the network's structure remains the same and hence this can be supported by existing deep learning libraries.
They proposed pruning as an optimization problem and their approach depends on statistics collected from the next layers and not the current layer.

\section{Proposed Method}

\subsection{Problem Formulation}
We restrict our analysis to the task of supervised classification of  3-dimensional synthetic datasets using Multi Layer Perceptron(MLP ) and classification of images using Convolutional Neural Network.
The classification task can be viewed as a many-to-one mapping, $f:R^d \mapsto \{c_1, c_2, \ldots c_k\}$.
The set of all samples  form a point cloud dataset on $d$-dimensional space ( $d=H \times W $  in the case of an image of height $H$  and width $W$ ).
As detailed in \cite{naitzat2020topology},  for classification task a non-homeomorphic transfer function is required for each layer as it can change the topology of point cloud dataset.
In order to achieve classification, one needs to change the topology from an initial complex topology to a simple and contractible topology for each class.
Another important characteristic of layer transfer function is that each layer reduces the dimensionality of the input data.

Our study focuses on two important aspects of neural network design.
\begin{enumerate}
\item Design of an optimum non-linear activation function for supervised classification task. See subsection \ref{design_of_activation_function}
\item Propose an easy technique for neural network pruning (i.e reducing the parameters in a trained model without significant decrease in accuracy) using betti numbers computed on the output feature space of each layer. See subsection \ref{network_pruning_1} for details
\end{enumerate}
\subsection{Dataset}

\begin{figure}[ht]
\begin{subfigure}{.5\textwidth}
  \centering
  % include first image
  \includegraphics[width=.5\linewidth]{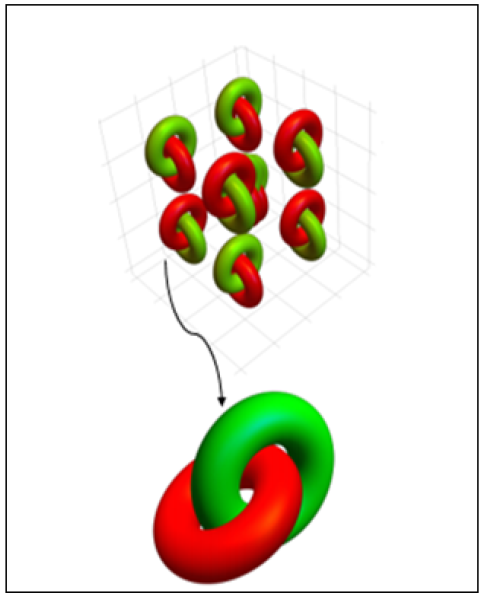}
  \caption{Nine ring dataset}
  \label{ninering}
\end{subfigure}
\begin{subfigure}{.5\textwidth}
  \centering
  % include second image
  \includegraphics[width=.48\linewidth]{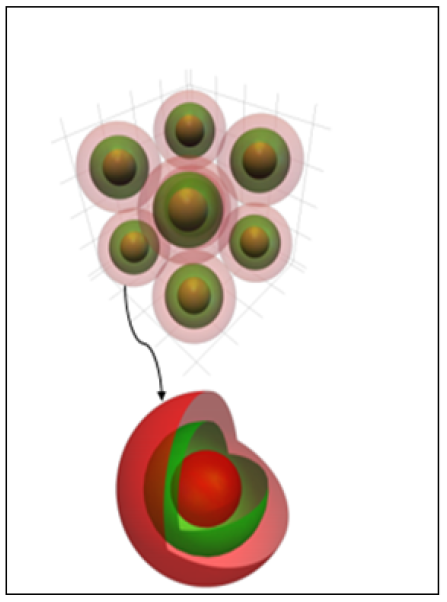}
  \caption{Nine sphere dataset}
  \label{ninesphere}
\end{subfigure}
\vspace{.1in}
\caption*{source:https://arxiv.org/pdf/2004.06093.pdf}
\vspace{.1in}
\caption{3-dimensional synthetic datasets with two different classes(Red and Green) used for training MLP}

\label{nine}
\end{figure}

We use two 3-dimensional simulated datasets, nine ring dataset and nine sphere dataset used in \cite{naitzat2020topology}.
The  nine ring dataset, as shown in Figure \ref{ninering},  consists of two classes of data,  colored Green and Red, interlocked together.
The nine sphere dataset consists of nine Green spheres and 18 Red Spheres enclosing each other as shown in Figure \ref{ninesphere}.
Both the datasets contain 16000 samples for training and 2000 samples for testing.

We also performed experiments and provide results on the following real datasets:
\begin{enumerate}
\item Cat-Dog dataset in Kaggle - The dataset consists of  images of size $32 \times 32$. The dataset is divided into training and testing sets, with 8000 training images and 2000 testing images,  each set with an equal number of images of cats and dogs.
\item Fashion Minst - The dataset consists of 70,000 grayscale images of size $28 \times 28$ with 10 different categories. The dataset is divided into training and testing sets, with 60,000 training images and 10,000 test images.
\item CIFAR-10 - The dataset consists of 60,000 colour images of size $32 \times 32$ with 10 different categories. Each category consists of 6000 images. The dataset is divided into training and testing sets, with 50,000 training images and 10,000 test images.
\end{enumerate}

Some sample images from each of the above datasets are shown in Figure \ref{catdog} , Figure \ref{fashionmnist}  and Figure \ref{cifar10}.
The datasets were converted to gray scale and normalized. No other preprocessing was performed.

\subsection{Design of Layer Transfer Function}
\begin{figure}
    \centering
    \includegraphics[width=.9\linewidth]{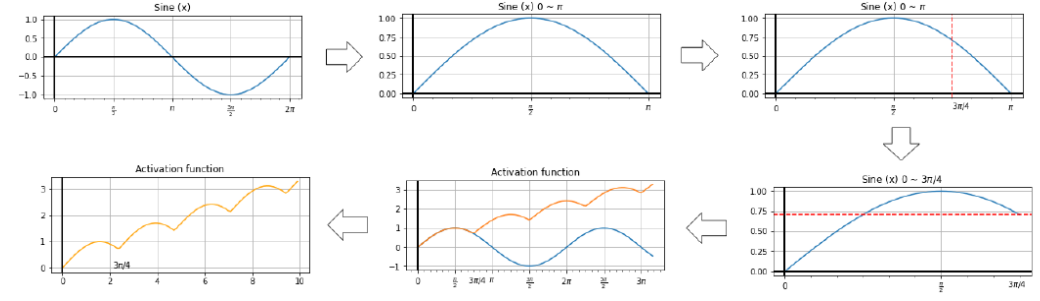}
        \vspace{.1in}
    \caption{Steps in design of many-to-one activation function}
    \label{figure11}
    \end{figure}

The layer transfer function is composed of an affine transformation and a non-linear activation function.
In the subsections below, we provide the details of desired characteristics for activation function from a topological point of view.

\subsection{Betti numbers and their significance on  layer transfer function}
\begin{figure}
    \centering
    \includegraphics[width=.7\linewidth]{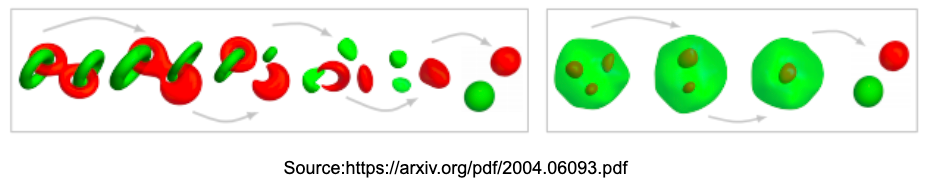}
\vspace{.1in}
    \caption{Simplification of topological space as the data is transformed by successive layers of neural network}
    \label{figure6}
    \end{figure}

Betti numbers, denoted as $\beta_k(X)$,   are used to quantify the topological complexity of a d-dimensional topological space  $X$, where  $0 \leq k \leq d$.
The  $0^{th}$ betty number, $\beta_0(X)$, is the number of connected components, the first betti number, $\beta_1(X)$, is the number of one dimensional holes, the second betti number is the number of two dimensional holes and so on.
%Figure \ref{figure5} shows some topological spaces and their corresponding betti numbers.
For efficient classification, one needs to transform the original point data cloud, $X$,  from a high dimensional space with large betti numbers to a low dimensional latent representation with  $\beta_0(X)$ (number of connected components) equal to the number of classes $K$ and all other betti numbers to zero.
As evident from Figure \ref{figure6} , this ensures that each connected component  corresponds to a single class ( either Red or Green) and there are no holes within the connected components.
Hence  each connected component is contractible  to a single point.
It is easy to find a decision boundary if there are no holes on the manifold formed by samples from a  single class.
In the rest of this document wherever we mention the term  topological complexity, we mean the betti numbers of the topological space.

%\begin{figure}
%\centering
%\includegraphics[width=.7\linewidth]{images/figure5.png}
%\vspace{0.1in}
%\caption{Examples of some complex topological spaces and corresponding Betti numbers}
%    \label{figure5}
%\end{figure}

\subsubsection{Design of activation function}\label{design_of_activation_function}
It is observed that non-homeomorphic activation functions like ReLU reduces the betti numbers sharply as opposed to traditional  activation functions like sigmoid and tanh which are  homeomorphic \cite{naitzat2020topology}.
Further, the more the number of discontinuities, the more powerful the activation function will be in terms of reducing the topological complexity.
In addition to these findings, we also hypothesize that multiple many-to-one regions in the layer transfer function can reduce the topological complexity  of samples within a single class, as it tries to bring more samples together.
As shown in Figure \ref{figure11}, we start with  a portion of a single half cycle of a sine function, and select a cut-off point $x=\frac{3\pi}{4}$.
The selected portion of the sine function is superimposed on a ReLU function as shown in the last figure in Figure \ref{figure11}.

The final analytical form of activation function is
\begin{equation}
y = ksin(\frac{3\pi}{4})   + sin(x - \frac{3\pi}{4}))
\end{equation}
%\vspace{0.2in}
 where $k=\left \lfloor{\frac{x}{\frac{3\pi}{4}}}\right \rfloor$
%\vspace{0.2in}
\subsection{Neural Network Architecture}
\begin{figure}
    \centering
    \includegraphics[width=.6\linewidth]{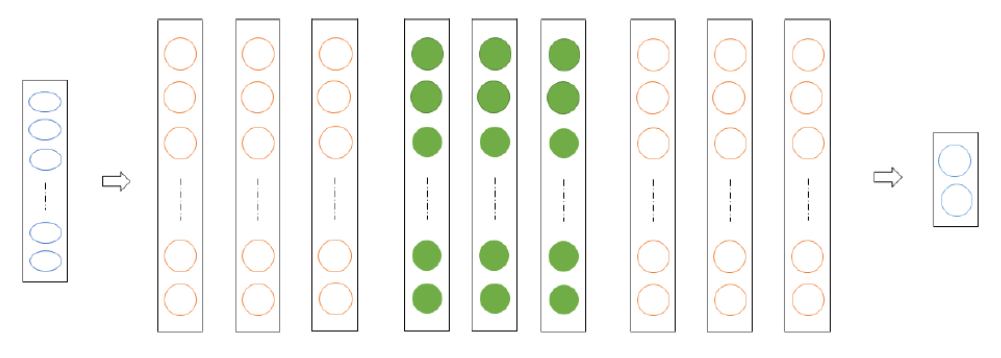}
    \vspace{.1in}
    \caption{Architecture of Multi layer perceptron used for training nine-ring and nine-sphere datasets. 9 hidden layers with 25 neurons in each layer. Proposed custom action function added in few of the hidden layers in the middle }
    \label{figure8}
    \end{figure}

\begin{figure}
    \centering
    \includegraphics[width=.7\linewidth]{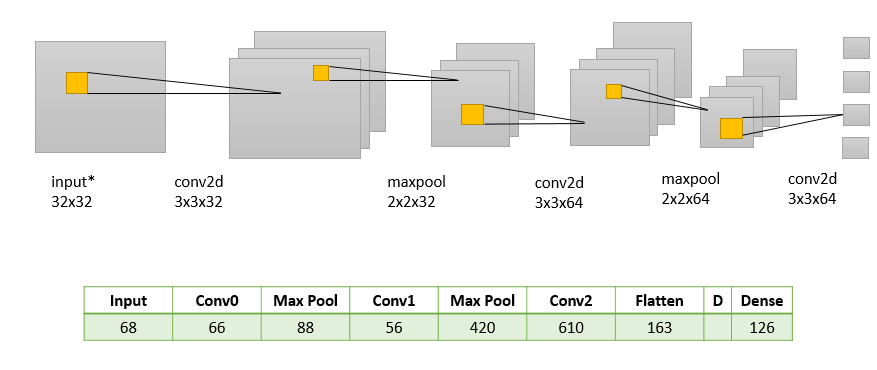}
    \caption{Architecture of Convolutional Neural Network for classifying images in cifar10 dataset along with the progression of betti numbers across the layers using ReLU activation function in all hidden layers}
    \label{figure9}
    \end{figure}

\begin{figure}
    \centering
    \includegraphics[width=.7\linewidth]{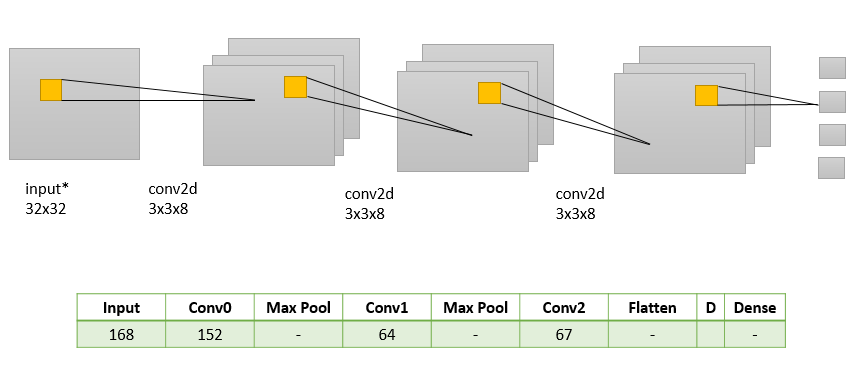}
    \caption{Architecture of Convolutional Neural Network for classifying images in cat vs dog dataset along with the progression of betti numbers across the layers using proposed activation function in few of the hidden layers}
    \label{figure10}
    \end{figure}

We performed experiments using MLP  on nine sphere and nine rings datasets.
The network architecture for MLP is shown in Figure \ref{figure8} . For image datasets, we used CNN shown in Figure \ref{figure9} and Figure \ref{figure10}.

\begin{figure}
    \centering
    \includegraphics[width=.7\linewidth]{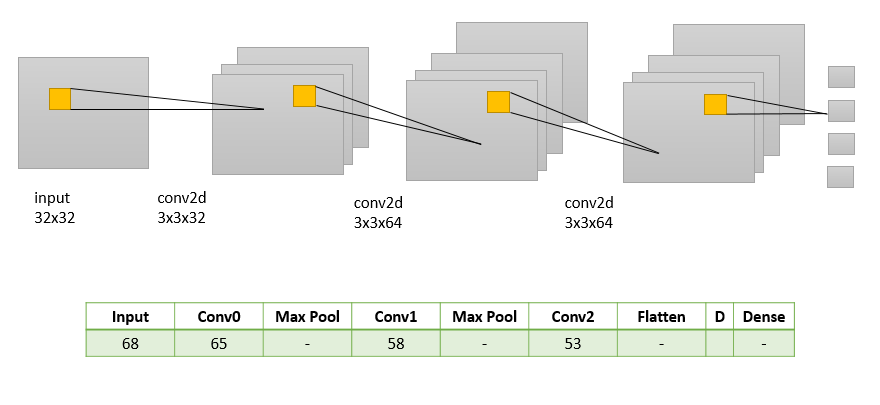}
    \caption{Architecture of CNN for classifying cifar10 images with proposed activation function added in few layers. It is observed that Betti numbers reduces significantly as compared to Figure \ref{figure9}  }
    \label{figure25}
    \end{figure}

The MLP for simulated dataset consists of 9 layers with 25 neurons in each layer.
We performed experiments with different legacy activation functions such as LeakyRelu (with different leak values), tanh, sigmoid and compared the results. Based on our findings on these results we proposed a new custom activation function and our empirical results shows that the proposed activation function performs better than existing ones.

\subsection{Pruning of Convolutional Neural Network}\label{network_pruning_1}
Convolutional neural networks have multiple (ranging from 10 - 1000 based on the complexity of task) channels or feature maps in the hidden layers. After training, it is possible that only a few subset of these feature maps are contributing significantly for the classification task. Other feature maps can be removed from the network without significant reduction in classification accuracy. This can result in reduced memory requirement and higher speed of prediction.

We propose a novel technique for identifying the significant  feature maps. We use the Betti number of the feature space to decide whether the feature map is significant or not.  Kernels that produce very large betti numbers are excluded.  Large betti numbers try to scatter the data points within the same class and hence increase intra-class distance between samples .

\section{Results and Discussion}
\begin{figure}[ht]
\begin{subfigure}{.5\textwidth}
  \centering
  % include first image
  \includegraphics[width=.8\linewidth]{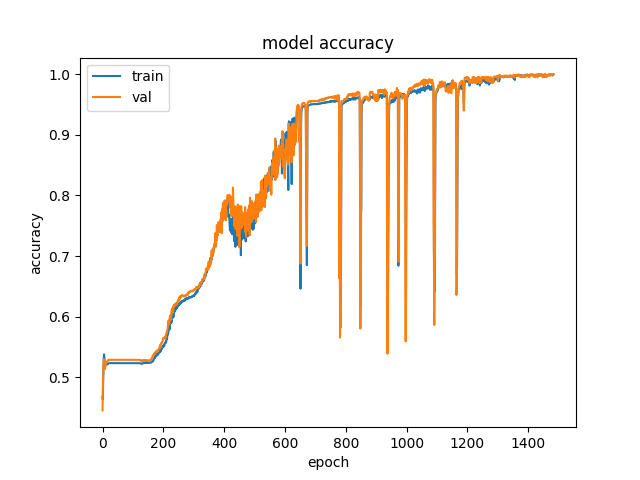}
  \caption{Convergence using proposed activation function}
  \label{custom}
\end{subfigure}
\begin{subfigure}{.5\textwidth}
  \centering
  % include second image
  \includegraphics[width=.8\linewidth]{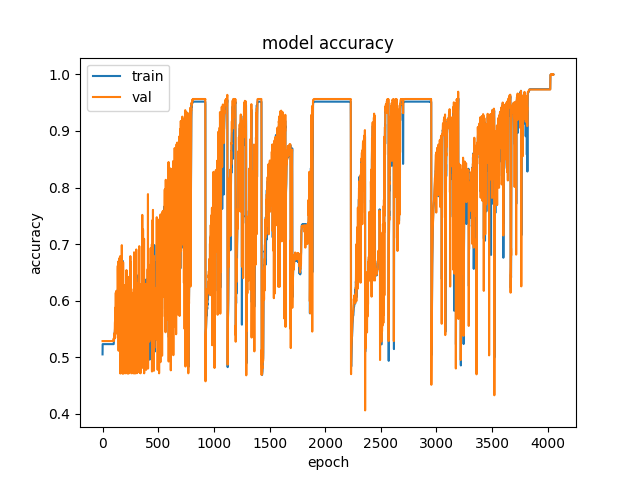}
  \caption{Convergence using ReLU activation function}
  \label{legacy}
\end{subfigure}
\vspace{0.1in}
\caption{Comparison of training convergence using proposed activation function and ReLU activation function. It is observed that convergence is faster with the propsed activation function}
\label{accuracy}
\end{figure}

It is observed, from Figure \ref{figure9} and \ref{figure25},   that with the proposed activation function Betti numbers decrease  by a significantly larger amount than with legacy activation functions.
Figure \ref{accuracy} shows the comparison of convergence  of multi layer perceptron with legacy activation functions and proposed activation function using nine-sphere and nine-ring datasets.
As evident from this figure, the reduction in betti numbers  directly translates to faster training convergence.
The new activation functions with many-to-one regions seem to work well on various datasets and help in achieving at-par or better trainability.

Also, The performance impact becomes more pronounced at larger batch-sizes i.e. with larger batch-sizes the gain tends to reach a factor of 2. On increasing the batch size the training for even simpler datasets takes longer (more epochs) for legacy activation functions. Even though the increase in training epochs  is seen for the proposed activation function also, the increase is less pronounced.

We further observed that,  in contrast to legacy activation functions like Relu etc. the need to adjust learning rate seems to be little. Hence the tuning hyper parameters like batch

Experiments were carried out on the MNIST dataset  to evaluate the impact of the proposed activation function on generalizability of the trained model. Both the legacy networks and custom networks achieve 96\% training and testing accuracy.
Only when the number of neurons in each layer were increased exponentially (to 500), did both the networks reach a training accuracy of 100\%. The testing accuracy remained at 96\% in both the cases, which shows that the proposed activation function didn’t add anything extra to overfitting.

\section{Conclusion}
In this work, we look at topological complexity of data at the output of each layer of deep neural network for binary classification tasks. Our contributions in this paper are two-fold: 1) design a new activation function that simplifies the topological complexity of point cloud data ( measured using Betti numbers)  at each hidden layer which translates to faster training convergence. 2) use Betti numbers  of feature space as a measure  for identifying and  pruning insignificant filters from a trained model hence making the model compact and faster prediction. We evaluate the proposed methods on popular image classification datasets and report results.

\bibliographystyle{ieeetr}
\bibliography{ms}

\end{document}